\documentclass{article}

\usepackage{amsmath,amsfonts,amssymb,times,graphicx,natbib,algorithm,algorithmic}
\usepackage{times}
\usepackage{graphicx} 
\usepackage{subfigure} 

\usepackage{natbib}

\usepackage{algorithm}
\usepackage{algorithmic}
\usepackage[usenames, dvipsnames]{color}

\usepackage[breaklinks=true]{hyperref}
\usepackage{breakurl}

\newcommand{\joschi}[1]{#1}

\usepackage[accepted]{whi2016}

\icmltitlerunning{Using Visual Analytics to Interpret Predictive Machine Learning Models}

\begin{document} 

\twocolumn[
\icmltitle{Using Visual Analytics to Interpret Predictive Machine Learning Models}

\icmlauthor{Josua Krause}{josua.krause@nyu.edu}
\icmladdress{New York University Tandon School of Engineering,
            Brooklyn, NY 11201 USA} 
\icmlauthor{Adam Perer}{adam.perer@us.ibm.com}
\icmladdress{IBM T.J. Watson Research Center,
                        Yorktown Heights, NY 10598 USA}
\icmlauthor{Enrico Bertini}{enrico.bertini@nyu.edu}
\icmladdress{New York University Tandon School of Engineering,
            Brooklyn, NY 11201 USA} 
\icmlkeywords{interpretability, predictive models, visual analytics, visualization, machine learning}

\vskip 0.3in
]

\begin{abstract} 
It is commonly believed that increasing the interpretability of a machine learning model may decrease its predictive power.  However, inspecting input-output relationships of those models using visual analytics, while treating them as black-box, can help to understand the reasoning behind outcomes without sacrificing predictive quality.  We identify a space of possible solutions and provide two examples of where such techniques have been successfully used in practice. 
\end{abstract} 

\section{Introduction}

There is growing demand among data scientists to generate and deploy predictive models in a variety of domains so that the patterns unearthed from massive amounts of data can be leveraged and converted into actionable insights. Predictive modeling is defined as the process of developing a mathematical tool or model that generates an accurate prediction \cite{kuhn2013applied}. Data scientists often turn to machine learning, where the goal is to create predictive models based on information automatically learned from data with ground truth. However, these machine learning techniques are often black-boxes and may be selected based only on performance metrics such as high accuracy scores, and not necessarily based on the interpretability and actionable insights of the model. 

Recently, we have worked on a variety of techniques to make predictive models more interpretable by bringing humans-in-the-loop through visual analytics. In this paper, we provide initial reflections on interpretability and the role visual analytics plays in this space based on our experience building such solutions. We first reflect on the role and meaning of interpretation and model transparency, then we reflect on the role of visual analytics in this space, then we describe two visual analytics systems we developed for model interpretation, and finally we conclude by proposing interesting questions for further research.



\section{Why and when is interpretation needed?}

It is important to start this discussion by clarifying that interpretation may not always be necessary in machine learning. There are plenty of situations in which building a model, testing and refining it, and finally putting it in production is absolutely appropriate (\emph{e.g.}, chess playing or face recognition). It is also important to point out that interpretation is necessarily a human activity and, as such, it may be costly and error prone. When, then, do we need human interpretation? Why do we need to involve humans in the machine learning process?  Without loss of generality, we have identified three main opportunities/needs in which interpretability is a highly desirable feature:
\vspace*{-0.5em}
\begin{enumerate}
    \item \textbf{Data understanding and discovery.} Machine learning is typically used as a tool to make predictions but it does not have to be used exclusively for this purpose. Machine learning models can also be used as a way to help understand and observe complex realities in providing abstractions that can be used by humans to enable interpretation and discovery.
    \item \textbf{Trust building and accountability.} Some mission-critical situations, e.g., when models make important decisions about human beings, it is important to have a better understanding of what a model does and why. Increasing model interpretation and transparency can play a role in increasing trust and accountability. 
    \item \textbf{Model comparison and diagnostics.} Model developers often need to understand where models fail to make correct decisions and also how multiple models differ. This is another circumstance in which interpretability can play a positive role. 
\end{enumerate}
\vspace*{-0.5em}
It is important to notice that many subjective properties play a role in model interpretation: \emph{e.g.}, plausibility, comprehension, trustworthiness, actionability.

\begin{figure}
\centering
\includegraphics[height=17.2em]{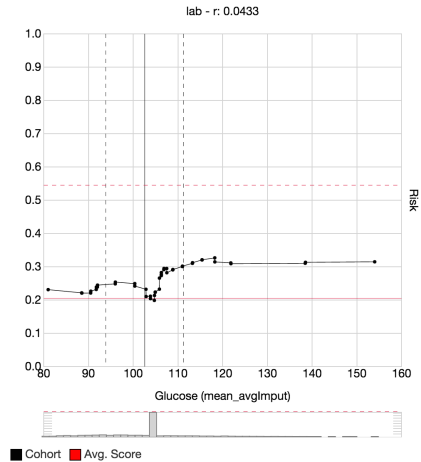} 
\caption{
Debugging model performance using partial dependence.
Instead of a direct relationship between higher Glucose lab values (x-axis) and higher risk scores (y-axis)
the model predicts a low risk for average Glucose lab values.
\joschi{The histogram below the plot, showing the distribution of the values found in the input data, indicates that most patients have the average value.}
Since missing values are imputed using the average Glucose value the valley in the plot
can be explained by the outcome independence of this value due to the high number of missing values.
}
\label{figs:pdp}
\end{figure}

\section{Model Transparency, Representation, and Interpretability}

While a machine learning technique is often defined as being more or less interpretable, it is important to point out that it is hard to \joschi{assign a} level of interpretability \joschi{to} a whole class of techniques. For instance, \textit{classification rules} are considered as a highly interpretable method but: how does one interpret the results of classification rules when the number of rules is very high (e.g., in the order of hundred or even thousands)? Similarly, \textit{neural networks}, are commonly regarded as having low interpretability, yet, recent visualization methods allows to look into some of the internal decisions the network makes which increase the interpretability of the model~\cite{yosinski-2015-ICML-DL-understanding-neural-networks,DBLP:journals/corr/XuBKCCSZB15,DBLP:journals/corr/ZeilerF13}.

We therefore propose to make the following distinctions. First, we propose to distinguish between \textit{model structure} and \textit{model representation}. As we will explain with our examples, it is possible through visual analytics to represent a model even by not having access to its internal logic or structure. Therefore, we propose that its interpretability cannot be defined exclusively by what specific training method was used or what internal representation/structure has been derived. Second, we propose to describe models as having different degrees of transparency, from fully opaque to fully transparent and define their representations as more or less interpretable.

This observation is particularly important when we consider the commonly held belief that interpretability and accuracy in machine learning are at odds~\cite{breiman2001}. While we cannot definitely refute such statement here, we believe it is important to consider that this may actually be a false dichotomy. If models can be interpreted by looking at input/output behavior, disregarding what internal structures produce such behavior, then it is possible to reconcile the need of using highly accurate models and yet retain interpretability.

\begin{figure}[t]
\centering
\includegraphics[height=17.2em]{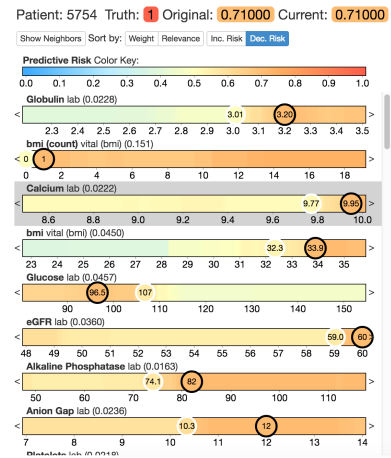} 
\caption{
Identifying changes to features that reduce the risk of a high risk patient.
The features are sorted by decreasing impact \joschi{to make large sets of features manageable}.
The background color of each feature indicates the predicted risk for this value.
\joschi{Impossible values, if known, appear grayed out and the slider snaps back to the last possible value if selected.}
}
\label{figs:dec_risk}
\end{figure}
\begin{figure*}[ht!]
\centering
\includegraphics[height=17.8em]{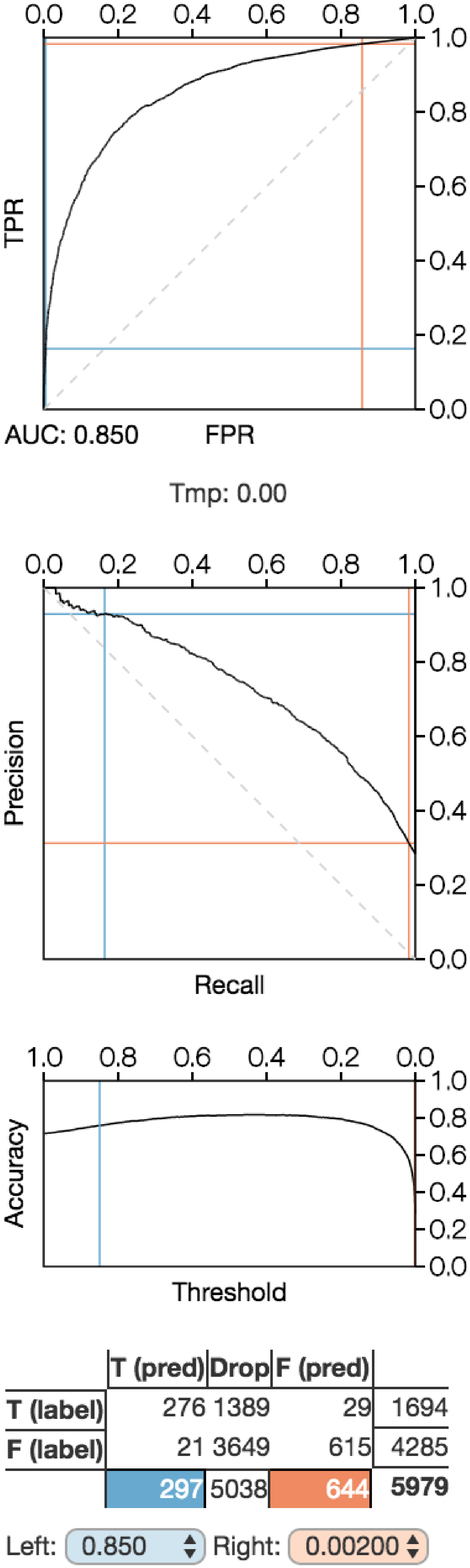} 
\includegraphics[height=17.8em]{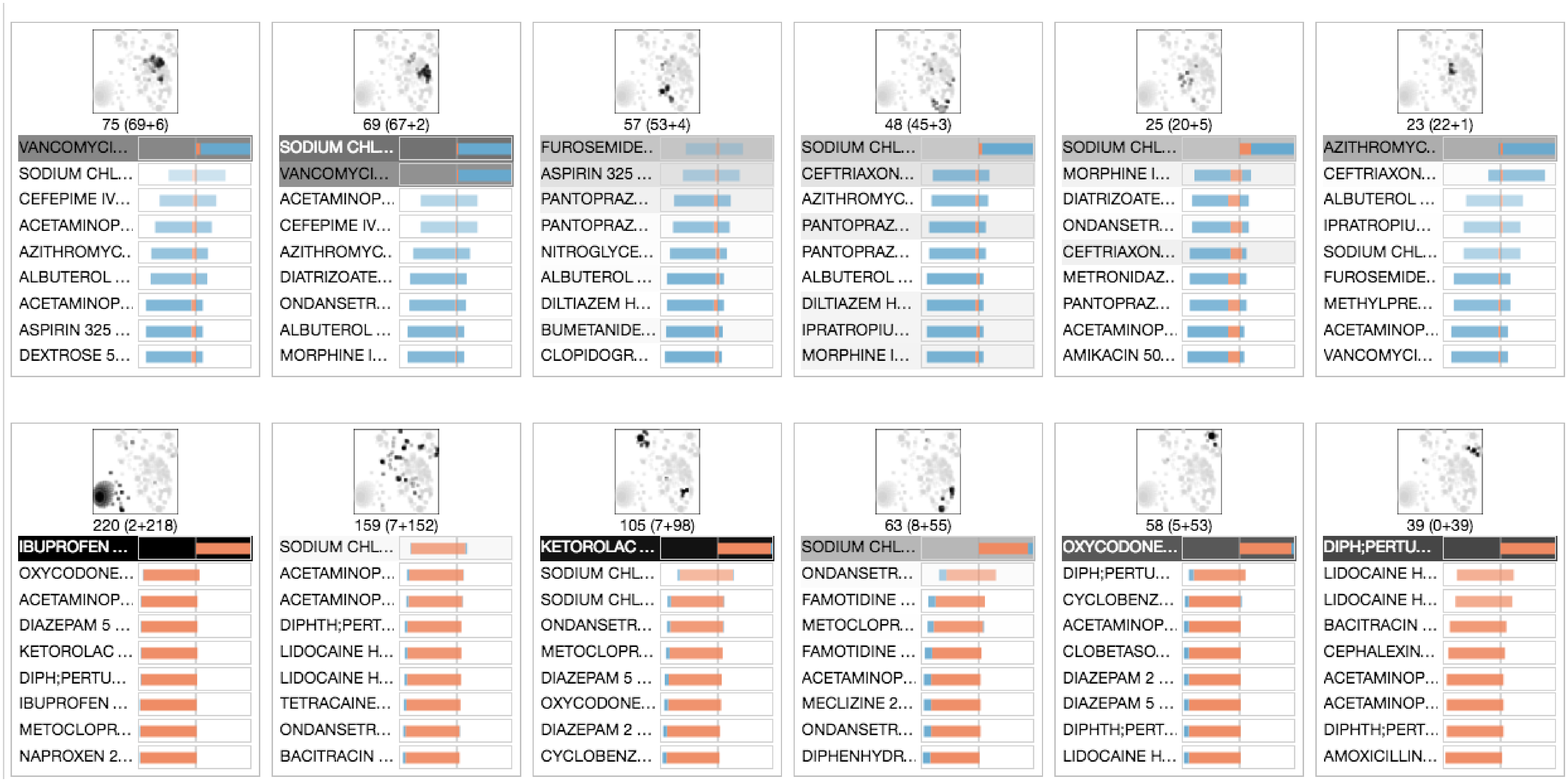} 
\caption{
Using class signatures to describe groups of patients admitted to the hospital
\joschi{because of} different \joschi{medications}.
Each column represents one group \joschi{(\emph{cluster}-step)} whereas each row shows the amount of patients in this group taking a
particular medication (the bar from the middle towards the right shows the percentage of patients taking the medication; the bar towards the left shows not taking medication).
The color of the bar shows the distribution of the true outcome labels as found in the input data.
The background of the rows shows the discriminativeness of a medication (dark being more discriminative \emph{wrt.} all other clusters\joschi{; \emph{rank}-step}).
\joschi{Above each group a \emph{t-SNE} projection of the items shows its relation to the other groups.}
ROC\joschi{,} precision-recall\joschi{, and accuracy} curves are shown on the left to facilitate selecting two thresholds to filter high signal patients \joschi{(\emph{contrast}-step)}.
}
\label{figs:class_signatures}
\end{figure*}

\section{The Role of Visual Analytics in Interpretation}

Visual analytics uses the power of visual perception as a way to involve humans in solving complex problems with data. Such human involvement, broadly speaking, can be desirable for two reasons: (1) because humans are better than machines in solving some tasks; (2) because human understanding and interpretation is desired. By designing perceptually effective visual representations~\cite{munzner14,Treisman:1985:PPV:5088.5091,Ware:2004:IVP:983611} we can enable fast, accurate, and hopefully trustworthy interpretation of machine learning models.

We identify two generic modalities that can be used for model interpretation with visual analytics:

\begin{enumerate}
    \item \textbf{Visualizing Model Structure (White-Box).} For transparent models, e.g., \textit{decision trees}, one option is to use visualization to represent the structure built through the training method. Several examples exist in this area, especially for \textit{decision trees} and \textit{rules} (\emph{eg.,}~\cite{6102453,yang2005pruning,liu2006rule}).
    \item \textbf{Visualizing Model Behavior (Black-Box).} Another option is to use visualization as a way to look at the \textit{behavior} of the model by looking exclusively at the relationship between input and output.
\end{enumerate}

Following, we focus exclusively on the second case.

\subsection{Visualizing Model Behavior (Input/Output)}

Model interpretation through visualization of input/output behavior of a model has a number of interesting properties, and it has received so far\joschi{,} comparatively, less attention than white-box approaches.

The most interesting property of this approach is that it does not depend on what specific method has been used to train a model and, as such, it has the big advantage of being extremely flexible and generic. Furthermore, by creating methods that enable interpretation of models by looking exclusively at their behavior, we can study the role of interpretation independently from model representation.

We have identified three main mechanisms through which model behavior can be observed and analyzed:

\begin{enumerate}
    \item \textbf{Item(s) to outcome.} In this modality, the user ``probes" the model by creating input examples with desired properties and observing what output the model generates (e.g., an input image obtained from a camera or a patient descriptor obtained by specifying a given set of properties).
    \item \textbf{Single feature to outcome.} In this modality, the user inspects and observes the relationship between one feature at a time and how its values relate (correlate) to the outcome. In our first example below (\textbf{Prospector}), we show how this can be done in practice.
    \item \textbf{Multiple features to outcome.} Finally, in this modality, the visualization aims at representing the relationship between many features and their values and how they related to the outcome. In our second example below (\textbf{Class Signatures}), we show one way this idea can be realized in practice.
\end{enumerate}

An additional aspect worth mentioning in relation to these three mechanisms, is how the data necessary to observe input/output behavior is obtained. Here, we have identified three main options (not mutually exclusive): \textit{training data}, \textit{test data (hold-out)}, \textit{simulated data}.

While the first two types of data are very common in machine learning training and validation steps, we notice that simulated data is much less common.

With simulated data we mean data that is synthetically generated (and as such may not belong to training or test data) by letting the user specify its properties.  One example of this situation is given in the example below (\textbf{Prospector}), in which, in a disease prediction task based on electronic health records, the user can specify a ``fictional" patient by dragging sliders that define his or her values.

Being able to work with \textit{simulated data} seems to be a particularly useful and promising direction of research when we consider the idea of probing machine learning models through visual analytics.

\section{Prospector}
\textbf{Prospector} is a novel visual analytics system designed to help analysts better understand predictive models \cite{prospector16}.  \textbf{Prospector} aims to support data scientists to go beyond judging predictive models solely based on their accuracy scores by also including model interpretability and actionable insights. It leverages the concept of partial dependence \cite{friedman2001}, a diagnostic technique that was designed to communicate how features affect the prediction, and makes this technique fully interactive. 

Figure~\ref{figs:pdp} shows how partial dependence can be used to debug machine learning models in \textbf{Prospector}. In this example imputation of missing values created unexpected behaviour of the inspected
classifier. \joschi{Partial dependence is given by}
\joschi{
\[
pdp_f(v) = \frac{1}{N} \sum_i^N pred(x_i) \;\text{with}\; x_{if} = v
\]
}
\joschi{
where $N$ is the number of rows in the input matrix $x$,
$pred$ is the prediction function that takes one input row, a feature vector, and returns a prediction score,
and $f$ is the feature used to compute the partial dependence plot.
The formula computes the average outcome over all input rows while changing the value of feature $f$ to the
input value $v$ for each row $x_i$. The original input data is kept fixed. This allows for observing the influence of $f$
on the prediction scores. Unlike generalized additive models, \emph{eg.,}~\cite{Caruana:2015:IMH:2783258.2788613}, this technique is model agnostic.}

\textbf{Prospector} also supports localized inspection, so users can understand why certain data results in a specific prediction, and even lets users hypothesize new data by tweaking values and seeing how the predictive model responds.  Users can interactively tweak feature values and see how the prediction responds, as well as find the most impactful features using a novel \joschi{model agnostic} local feature importance metric \joschi{that only depends on partial dependence}.

Figure~\ref{figs:dec_risk} shows the prediction inspection portion of the \textbf{Prospector} UI, which allows users to examine the features contributing to the prediction of a selected data point. All of the features’ partial dependence bars are shown in a scrollable list, with the data's feature values set with circular labels. Users can drag the circular label to change the value of any feature and see the prediction change in real-time.
Users can change the sort order of the partial dependence bars by using the buttons at the top. In addition to sorting by the feature weight and relevance as determined by the predictive model \joschi{if available}, users can also sort according to our local feature importance and impactful changes. If impactful changes are chosen as the sort order, the suggested changes to each feature are indicated with a white circular label in the partial dependence bar.

\begin{figure}[t]
\centering
\includegraphics[width=0.49\linewidth]{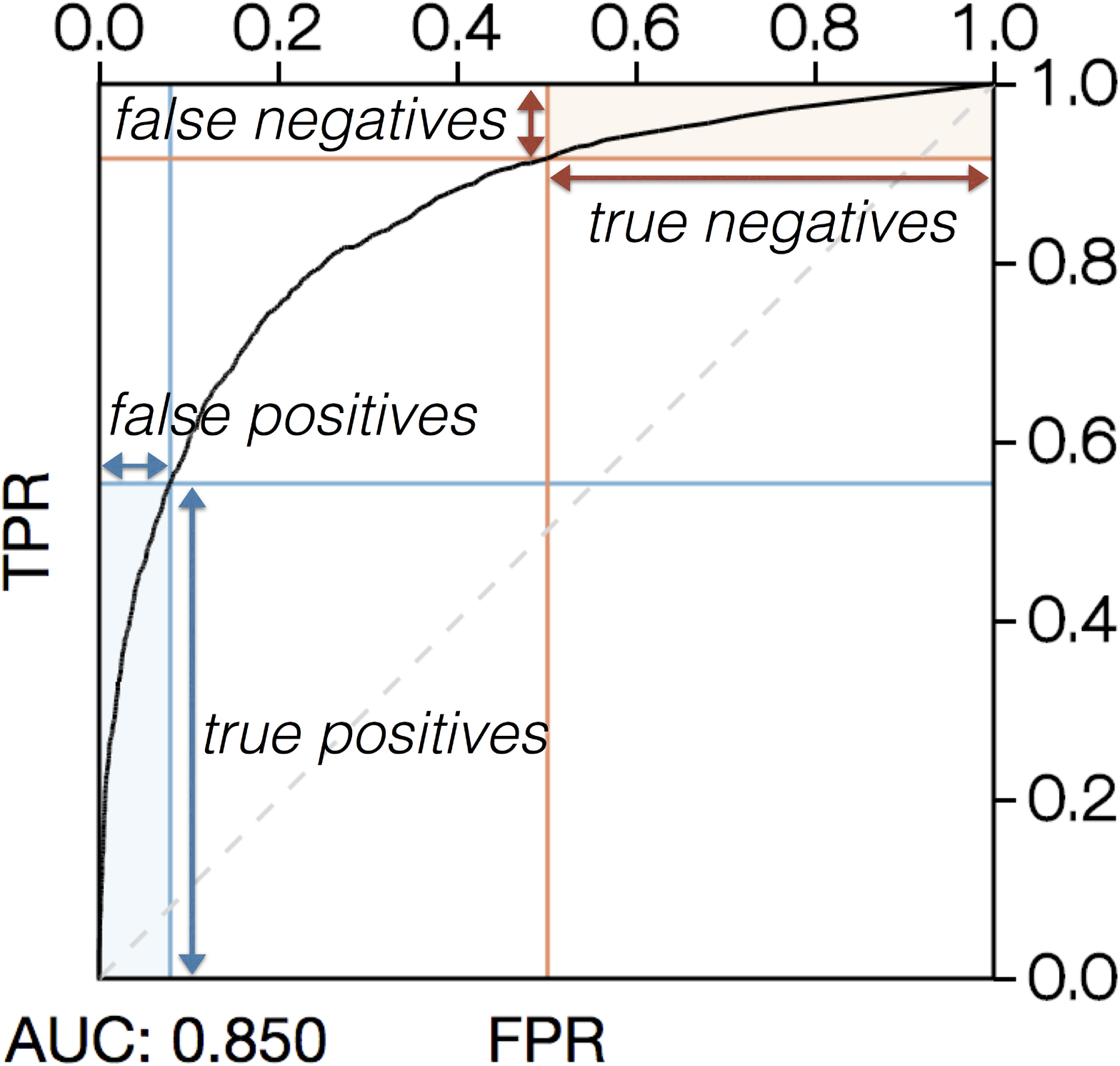} 
\includegraphics[width=0.42\linewidth]{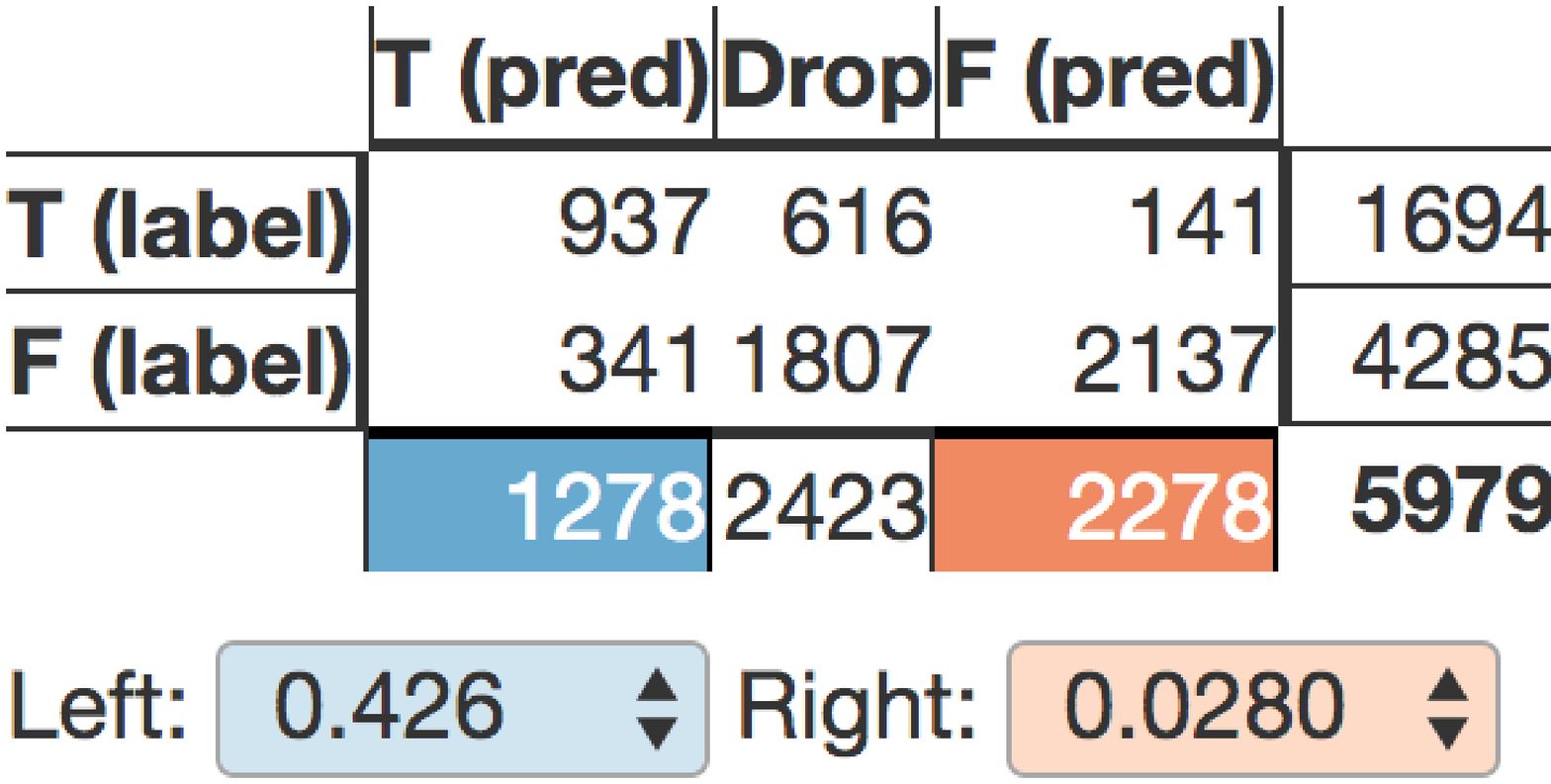} 
\caption{
\joschi{
Using the receiver operating characteristics curve (ROC-Curve) for determining filter thresholds.
The curve shows how the true positive rate (TPR; y-axis) and the false positive rate (FPR; x-axis)
changes with decreasing thresholds (the threshold values are not shown in the graph).
The extended contingency matrix on the right shows the actual numbers. 
}
}
\label{figs:roc_explain}
\end{figure}

\section{Class Signatures}

With \textbf{Class Signatures}, we propose a visual analytics workflow to interpret predictive associations between a large set of binary features and a binary target.
For this we use a $4$ step pipeline: \textit{model}, \textit{contrast}, \textit{cluster}, and \textit{rank}, and a visual analytics interface that allows the end user to detect and interpret such associations.
After modeling the predictive associations using a binary classifier we leverage the prediction scores with two user defined thresholds, one for positive cases and one for negative cases, to focus only on data items with a strong predictive signal, increasing contrast.
Then, we cluster both positive and negative examples \textit{separately}.
This groups together data points that have the same predicted outcome and a similar configuration of values.
Finally, we rank each feature in the computed clusters using discriminative analysis across \textit{all clusters}.

For interpreting the results with visual analytics, we use \textbf{Class Signatures} as shown in Figure~\ref{figs:class_signatures}.
As our input features are binary in nature we show in the class signatures how consistently present
a feature is in a cluster.
This is indicated by bars growing both to the right (percentage of which feature is present) and the left (percentage of which feature is not present).
In combination with the discriminative measure of the features (\joschi{computed as gini-importance;} the shade of feature backgrounds is visually encoded, so darker means more discriminative) users can formulate rules that explain predictions for different subgroups of data items.

The proposed workflow allows for a more fine-grained analysis of the driving factors of a predictive task than using commonly used feature importance techniques.
This is due to the observation that many phenomena have multiple underlying reasons for the same result.
Thus an explanation is needed that distinguishes which features were actually responsible for given data points.
\textbf{Class Signatures} provide this distinction in the form of user interpretable rules.

\section{Conclusion}
We have provided an initial characterization of how visual analytics can be used for model interpretation, with a focus on visualizing input/output behavior rather than model structure. To exemplify these approaches, we presented two practical examples of systems we built to understand the reasoning of classification algorithms while treating them as black-box. As such, these examples provide initial anecdotal evidence that we do not necessarily need to accept lower prediction performance in order to gain in interpretability (\joschi{Breiman} identifies an inverse relation between model performance and interpretability of a machine learning model \joschi{\cite{breiman2001}}). Of course, there is much future work to do as the space of possible solutions in analyzing and understanding input-output relationships of machine learning models using visual analytics is largely unexplored. We identify two major research trajectories to develop these ideas further: (1) develop and validate more solutions that exploit this specific modality and (2) study human interpretation through user studies aimed at better defining interpretation and establishing metrics useful for comparison of alternative solutions.

\bibliography{visualization}
\bibliographystyle{icml2016}

\end{document}